\documentclass[sigconf,screen]{acmart}

\usepackage{booktabs} 
\usepackage[justification=centering,font={footnotesize}]{caption}
\usepackage{amsmath}
\usepackage{lipsum}
\usepackage{balance}
\usepackage{algorithmic}

\setcopyright{acmlicensed}

\copyrightyear{2018}
\acmYear{2018}
\setcopyright{acmcopyright}
\acmConference[GECCO '18]{Genetic and Evolutionary Computation Conference}{July 15--19, 2018}{Kyoto, Japan}
\acmPrice{15.00}
\acmDOI{10.1145/3205455.3205578}
\acmISBN{978-1-4503-5618-3/18/07}

\begin{document}
\title{Evolving simple programs for playing Atari games}

\author{Dennis G Wilson}
\email{dennis.wilson@irit.fr}
\affiliation{
  \institution{University of Toulouse}
  \department{IRIT - CNRS - UMR5505}
  \city{Toulouse}
  \country{France}
  \postcode{31015}
}

\author{Sylvain Cussat-Blanc}
\email{sylvain.cussat-blanc@irit.fr}
\affiliation{
  \institution{University of Toulouse}
  \department{IRIT - CNRS - UMR5505}
  \city{Toulouse}
  \country{France}
  \postcode{31015}
}

\author{Herv\'e Luga}
\email{herve.luga@irit.fr}
\affiliation{
  \institution{University of Toulouse}
  \department{IRIT - CNRS - UMR5505}
  \city{Toulouse}
  \country{France}
  \postcode{31015}
}

\author{Julian F Miller}
\email{julian.miller@york.ac.uk}
\affiliation{
  \institution{University of York}
  \city{York}
  \country{UK}
}

\renewcommand{\shortauthors}{DG Wilson et al.}
\def\sectionautorefname{Section}
\def\subsectionautorefname{Section}

\begin{abstract}
  Cartesian Genetic Programming (CGP) has previously shown capabilities in image
processing tasks by evolving programs with a function set specialized for
computer vision. A similar approach can be applied to Atari playing. Programs
are evolved using mixed type CGP with a function set suited for matrix
operations, including image processing, but allowing for controller behavior to
emerge. While the programs are relatively small, many controllers are
competitive with state of the art methods for the Atari benchmark set and
require less training time. By evaluating the programs of the best evolved
individuals, simple but effective strategies can be found.

\end{abstract}

%
%

\ccsdesc[500]{Computing methodologies~Artificial intelligence}
\ccsdesc[300]{Computing methodologies~Model development and analysis}

\keywords{Games, Genetic programming, Image analysis, Artificial intelligence}

\maketitle

\section{Introduction}
\label{sec:intro}

The Arcade Learning Environment (ALE) \cite{bellemare13arcade} has recently been
used to compare many controller algorithms, from deep Q learning to
neuroevolution. This environment of Atari games offers a number of different
tasks with a common interface, understandable reward metrics, and an exciting
domain for study, while using relatively limited computation resources. It is no
wonder that this benchmark suite has seen wide adoption.

One of the difficulties across the Atari domain is using pure pixel input. While
the screen resolution is modest compared to modern game platforms, processing
this visual information is a challenging task for artificial agents. Object
representations and pixel reduction schemes have been used to condense this
information into a more palatable form for evolutionary controllers. Deep neural
network controllers have excelled here, benefiting from convolutional layers and
a history of application in computer vision.

Cartesian Genetic Programming (CGP) also has a rich history in computer vision,
albeit less so than deep learning. CGP-IP has capably created image filters for
denoising, object detection, and centroid determination. There has been less
study using CGP in reinforcement learning tasks, and this work represents the
first use of CGP as a game playing agent.

The ALE offers a quantitative comparison between CGP and other methods. Atari
game scores are directly compared to published results of multiple different
methods, providing a perspective on CGP's capability in comparison to other
methods in this domain.

CGP has unique advantages that make its application to the ALE interesting. By
using a fixed-length genome, small programs can be evolved and later read for
understanding. While the inner workings of a deep actor or evolved neural
network might be hard to discern, the programs CGP evolves can give insight into
strategies for playing the Atari games. Finally, by using a diverse function set
intended for matrix operations, CGP is able to perform comparably to humans on a
number of games using pixel input with no prior game knowledge.

This article is organized as follows. In the next section,
\autoref{sec:background}, a background overview of CGP is given, followed by a
history of its use in image processing. More background is provided concerning
the ALE in \autoref{sec:ale}. The details of the CGP implementation used in this
work are given in \autoref{sec:methods}, which also covers the application of
CGP to the ALE domain. In \autoref{sec:results}, CGP results from 61 Atari games
are compared to results from the literature and selected evolved programs are
examined. Finally, in \autoref{sec:discussion}, concerns from this experiment
and plans for future work are discussed.

\section{Background}
\label{sec:background}

While game playing in the ALE involves both image processing and reinforcement
learning techniques, research on these topics using CGP has not been equal.
There is a wealth of literature concerning image processing in CGP, but little
concerning reinforcement learning. Here, we therefore focus on the general
history of CGP and its application to image processing.

\subsection{Cartesian Genetic Programming}

Cartesian Genetic Programming \cite{Miller2000} is a form of Genetic Programming
in which programs are represented as directed, often acyclic graphs indexed by
Cartesian coordinates. Functional nodes, defined by a set of evolved genes,
connect to program inputs and to other functional nodes via their coordinates.
The outputs of the program are taken from any internal node or program input
based on evolved output coordinates.

In its original formulation, CGP nodes are arranged in a rectangular grid of $R$
rows and $C$ columns. Nodes are allowed to connect to any node from previous
columns based on a connectivity parameter $L$ which sets the number of columns
back a node can connect to; for example, if $L=1$, nodes could connect to the
previous column only. Many modern CGP implementations, including that used in
this work, use $R=1$, meaning that all nodes are arranged in a single row
\cite{Miller2011a}.

In recurrent CGP \cite{Turner2014} (RCGP), a recurrency parameter was introduced
to express the likelihood of creating a recurrent connection; when $r=0$,
standard CGP connections were maintained, but $r$ could be increased by the user
to create recurrent programs. This work uses a slight modification of the
meaning of $r$, but the idea remains the same.

In practice, only a small portion of the nodes described by a CGP chromosome
will be connected to its output program graph. These nodes which are used are
called ``active'' nodes here, whereas nodes that are not connected to the output
program graph are referred to as ``inactive'' or ``junk'' nodes. While these
nodes are not used in the final program, they have been shown to aid
evolutionary search \cite{Miller2006, Vassilev2000, Yu2001}.

The functions used by each node are chosen from a set of functions based on the
program's genes. The choice of functions to include in this set is an important
design decision in CGP. In this work, the function set is informed by MT-CGP
\cite{Harding2012} and CGP-IP \cite{Harding2013}. In MT-CGP, the function of a
node is overloaded based on the type of input it receives: vector functions are
applied to vector input and scalar functions are applied to scalar input. The
choice of function set is very important in CGP. In CGP-IP, the function set
contained a subset of the OpenCV image processing library and a variety of
vector operations.

\subsection{Image Processing}
\label{sec:image_proc}

CGP has been used extensively in image processing and filtering tasks. In
\citet{Montes2003}, centroids of objects in images were determined by CGP. A
similar task was more recently undertaken in \citet{Paris2015}, which detected
and filtered simple shapes and musical notes in images. Other image filters were
evolved in \citet{Smith2005} and \citet{Sekanina2011} which involved tasks such as
image denoising. Finally, \citet{Harding2008} demonstrated the ability to use
GPUs with CGP for improved performance in image processing tasks.

Many of these methods use direct pixel input to the evolved program. While
originally demonstrated using machine learning benchmarks, MT-CGP
\citep{Harding2012} offered an improvement to CGP allowing for greater image
processing techniques to follow. By using matrix inputs and functions, entire
images could be processed using state of the art image processing libraries. A
large subset of the OpenCV library was used in \citet{Harding2013} for image
processing, medical imaging, and object detection in robots.

\subsection{Arcade Learning Environment}
\label{sec:ale}

The ALE offers a related problem to image processing, but also demands
reinforcement learning capability, which has not been well studied with CGP.
Multiple neuroevolution approaches, including HyperNEAT, and CMA-ES were applied
to pixel and object representations of the Atari games in
\citet{hausknecht2014neuroevolution}. The performance of the evolved object-based
controllers demonstrated the difficulties of using raw pixel input; of the 61
games evaluated, controllers using pixel input performed the best for only 5
games. Deterministic random noise was also given as input and controllers using
this input performed the best for 7 games. This demonstrates the capability of
methods that learn to perform a sequence of actions unrelated to input from the
screen.

HyperNEAT was also used in \citet{hausknecht2012hyperneat} to show
generalization over the Freeway and Asterix games, using a visual processing
architecture to automatically find an object representation as inputs for the
neural network controller. The ability to generalize over multiple Atari games
was further demonstrated in \citet{kelly2017multi}, which followed
\citet{kelly2017emergent}. In this method, tangled problem graphs (TPG) use a
feature grid created from the original pixel input. When evolved on single
games, the performance on 20 games was impressive, rivaling human performance in
6 games and outperforming neuroevolution. This method generalized over sets of 3
games with little performance decrease.


The ALE is a popular benchmark suite for deep reinforcement learning. Originally
demonstrated with deep Q-learning in \citet{mnih2013playing}, the capabilities of
deep neural networks to learn action policies based on pixel input was fully
demonstrated in \citet{mnih2015human}. Finally, an actor-critic model improved
upon deep network performance in \citet{mnih2016asynchronous}.

\section{Methods}
\label{sec:methods}

While there are many examples of CGP use for image processing, these
implementations had to be modified for playing Atari games. Most importantly,
the input pixels must be processed by evolved programs to determine scalar
outputs, requiring the programs to reduce the input space. The methods following
were chosen to ensure comparability with other ALE results and to encourage the
evolution of competitive but simple programs. The code for this paper is
available as part of the CGP.jl
repository.\footnote{https://github.com/d9w/CGP.jl}

\begin{table}[t]
  \begin{tabular}{|c|c|c|c|}
    \hline
Function & Description & Arity & Broadcasting\\\hline
\multicolumn{4}{|c|}{Mathematical}\\\hline
ADD & $(x+y)/2$ & 2 & Yes\\
AMINUS & $|x-y|/2$ & 2 & Yes\\
MULT & $x y$ & 2 & Yes\\
CMULT & $x p_n$ & 1 & Yes\\
INV & $1/x$ & 1 & Yes\\
ABS & $|x|$ & 1 & Yes\\
SQRT & $\sqrt{|x|}$ & 1 & Yes\\
CPOW & $|x|^{p_n+1}$ & 1 & Yes\\
YPOW & $|x|^{|y|}$ & 2 & Yes\\
EXPX & $(e^{x}-1)/(e^1-1)$ & 1 & Yes\\
SINX & $\sin x$ & 1 & Yes\\
SQRTXY & $\sqrt{x^2+y^2}/\sqrt{2} $ & 2 & Yes\\
ACOS & $(\operatorname{arccos} x)/\pi$ & 1 & Yes\\
ASIN & $2(\operatorname{arcsin} x)/\pi$ & 1 & Yes\\
ATAN & $4(\operatorname{arctan} x)/\pi$ & 1 & Yes\\
\hline
\multicolumn{4}{|c|}{Statistical}\\\hline
STDDEV & $std(\vec{x})$ & 1 & No\\
SKEW & $skewness(\vec{x})$ & 1 & No\\
KURTOSIS & $kurtosis(\vec{x})$ & 1 & No\\
MEAN & $mean(\vec{x})$ & 1 & No\\
RANGE & $max(\vec{x})-min(\vec{x})-1$ & 1 & No\\
ROUND & $round(\vec{x})$ & 1 & No\\
CEIL & $ceil(\vec{x})$ & 1 & No\\
FLOOR & $floor(\vec{x})$ & 1 & No\\
MAX1 & $max(\vec{x})$ & 1 & No\\
MIN1 & $min(\vec{x})$ & 1 & No\\
\hline
\multicolumn{4}{|c|}{Comparison}\\\hline
LT & $x < y$ & 2 & Yes\\
GT & $x > y$ & 2 & Yes\\
MAX2 & $\operatorname{max}(x , y)$ & 2 & Yes\\
MIN2 & $\operatorname{min}(x , y)$ & 2 & Yes\\
\hline
  \end{tabular}
  \caption{A part of the function set used. Many of the mathematical and
    comparison functions are standard for inclusion in CGP function sets for
    scalar inputs. Where broadcast is indicated, the function was applied
    equally to scalar and matrix input, and where it is not, scalar inputs were
    passed directly to output and only matrix inputs were processed by the
    function.}
  \label{tab:functions}
\end{table}

\begin{table*}[t]
  \begin{tabular}{|c|c|c|c|}
    \hline
Function & Description & Arity & Broadcasting\\\hline
\multicolumn{4}{|c|}{List processing}\\\hline
SPLIT\_BEFORE & return all values before $\frac{p_n+1}{2}$ in $\vec{x}$ & 1 & No\\
SPLIT\_AFTER & return all values after $\frac{p_n+1}{2}$ in $\vec{x}$ & 1 & No\\
RANGE\_IN & return the values of $\vec{x}$ in $[\frac{y+1}{2}, \frac{p_n+1}{2}]$ & 2 & No\\
INDEX\_Y & return the value of $\vec{x}$ at $\frac{y+1}{2}$ & 2 & No\\
INDEX\_P & return the value of $\vec{x}$ at $\frac{p_n+1}{2}$ & 1 & No\\
VECTORIZE & return all values of $\vec{x}$ as a 1D vector & 1 & No\\
FIRST & return the first value of $\vec{x}$ & 1 & No\\
LAST & return the last value of $\vec{x}$ & 1 & No\\
DIFFERNCES & return the computational derivative of the 1D vector of $\vec{x}$ & 1 & No\\
AVG\_DIFFERENCES & return the mean of the DIFF function & 1 & No\\
ROTATE & perform a circular shift on $\vec{x}$ by $p_n$ elements & 1 & No\\
REVERSE & reverse $\vec{x}$ & 1 & No\\
PUSH\_BACK & create a new vector with all values of $x$ or $\vec{x}$, then $y$ or $\vec{y}$ & 2 & No\\
PUSH\_BACK & create a new vector with all values of $y$ or $\vec{y}$, then $x$ or $\vec{x}$ & 2 & No\\
SET & return the scalar value $x$ $\texttt{len}(\vec{y})$ times, or $y$ $\texttt{len}(\vec{x})$ & 2 & No\\
SUM & return the sum of $\vec{x}$ & 1 & No\\
TRANSPOSE & return the transpose of $\vec{x}$ & 1 & No\\
VECFROMDOUBLE & return the 1-element $\vec{x}$ if $x$ is a scalar & 1 & No\\
\hline
\multicolumn{4}{|c|}{MISCELLANEOUS}\\\hline
YWIRE & $y$ & 1 & No\\
NOP & $x$ & 1 & No\\
CONST & $p_n$ & 0 & No\\
CONSTVECTORD & return a matrix of $size(\vec{x})$ with values of $p_n$ & 1 & No\\
ZEROS & return a matrix of $size(\vec{x})$ with values of 0 & 1 & No\\
ONES & return a matrix of $size(\vec{x})$ with values of 1 & 1 & No\\
\hline
  \end{tabular}
  \caption{List processing and other functions in the function set. The choice
    of many of these functions was inspired by MT-CGP \cite{Harding2012}.}
  \label{tab:list_functions}
\end{table*}

\subsection{CGP genotype}

In this work, a floating point representation of CGP is used. It has some
similarity with a previous floating point representation~\cite{Clegg2007}. In
the genome, each node $n$ in $C$ columns is represented by four floats, which
are all bound between $[0.0, 1.0]$: $x$ input, $y$ input, function, parameter
$p$.

The $x$ and $y$ values are used to determine the inputs to $n$. The function
gene is cast to an integer and used to index into the list of available
functions, determining $f_n$. Finally, the parameter is scaled between $[-1.0,
  1.0]$ using $p_n = 2p -1$. Parameters are passed to functions, as they are
used by some functions. Parameters are also used in this work as weights on the
final function, which has been done in other CGP work \cite{Knezevic2017}.

Nodes are ordered based on their ordering in the genome. The genome begins with
$n_{output}$ nodes which determine the index of the output nodes in the graph,
and then all genes for the $C$ program nodes. The first $n_{input}$ nodes
correspond to the program inputs and are not evolved; the first node after these
will correspond to the first four floating point values after $n_{output}$ in
the genome, and the next node will correspond to the next four values, and so
on. The number of columns $C$ counts only the program nodes after $n_{input}$,
so, in total, the graph is composed of $N = n_{input} + C$ nodes and is based on
$G = n_{output} + 4C$ genes.

When determining the inputs for a node $n$, the $x_n$ and $y_n$ genes are scaled
according to $r$ and then rounded down to determine the index of the connected
nodes, $xi_n$ and $yi_n$. The value $r$ indicates the range over which $x_n$ and
$y_n$ operate; when $r=0$, connections are only possible between the first input
node and the $n$th graph node, and when $r=1$, connections are possible over the
entire genome.

\begin{gather*}
  xi_n = \lfloor x_n ((1-\frac{n}{N})r + \frac{n}{N}) \rfloor\\
  yi_n = \lfloor y_n ((1-\frac{n}{N})r + \frac{n}{N}) \rfloor
\end{gather*}

Output genes are also rounded down to determine the indices of the nodes which
will give the final program output. Once all genes have been converted into
nodes, the active graph is determined. Starting from the output nodes, $xi_n$
and $yi_n$ are used to recursively trace the node inputs back to the final
program input. Nodes are marked as active when passed, and nodes which have
already been marked active are not followed, allowing for a recursive search
over graphs with recurrent connections.

With the proper nodes marked as active, the program can be processed. Due to the
recurrent connections, the program must be computed in steps. Each node in the
graph has an output value, which is initially set to the scalar 0. At each step,
first the output values of the program input nodes are set to the program
inputs. Then, the function of each program node is computed once, using the outputs
from connected nodes of the previous step as inputs.

\begin{algorithmic}
  \FOR{$n=0$ to $n_{input}$} 
    \STATE $out_n = \texttt{program\_input[n]}$
  \ENDFOR
  \FOR{$n=n_{input}$ to $N$} 
    \STATE $out_n = p_n f_n(out_{xi_n}, out_{yi_n}, p_n)$
  \ENDFOR
\end{algorithmic}

The floating point representation in this work was chosen to simplify the genome
and evolution. It allows all genes to be represented as the same type, a float,
while still allowing for high precision in the evolution of the parameter gene.

\subsection{Evolution}

A standard $1+\lambda$ EA is used to evolve the programs. At initialization, a
random genome is created using $G$ uniformly distributed values in $[0.0, 1.0]$.
This individual is evaluated and is considered the first elite individual. At
each generation, $\lambda$ offspring are generated using genetic mutation. These
offspring are each evaluated and, if their fitness is greater than or equal to
that of the elite individual, they replace the elite. This process is repeated
until $n_{eval}$ individuals have been evaluated; in other words, for
$\frac{n_{eval}}{\lambda}$ generations. The stop condition is expressed here as
a number of evaluations to make runs comparable during optimization of
$\lambda$.

The genetic mutation operator randomly selects $m_{nodes}$ of the program node
genes and sets them to new random values, again drawn from a uniform random
distribution in $[0.0, 1.0]$. The output nodes are mutated according to a
different probability; $m_{output}$ of the output genes are randomly set to new
values during mutation. When these values have been optimized, they are often
found to be distinct. It therefore seems beneficial to include this second
parameter for output mutation rate.

\begin{table}[th]
  \begin{tabular}{l l | l l}
    $C$ & 40 & $m_{nodes}$ & 0.1\\
    $r$ & 0.1 & $m_{output}$ & 0.6\\
    $\lambda$ & 9 & $n_{eval}$ & 10000
  \end{tabular}
  \caption{CGP parameter values.\\All parameters except $n_{eval}$ were optimized
    using irace.}
  \label{tab:params}
\end{table}

The parameters $C$, $r$, $\lambda$, $m_{nodes}$, and $m_{output}$ were optimized
using irace \cite{lopez2016irace}. The values used in this experiment are
presented in \autoref{tab:params} and are somewhat standard for CGP. $\lambda$
is unusually large; normal values are 4 or 5, and the maximum allowed during
parameter optimization was 10. The other main parameter setting in CGP is the
choice of function set, which is detailed next.

\begin{figure*}[t]
  \centering
  \includegraphics[width=0.6\textwidth]{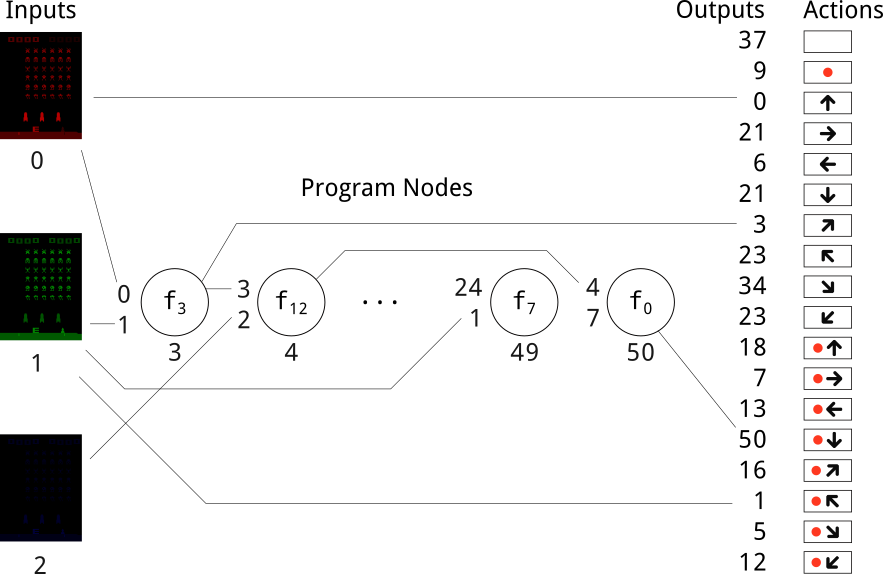}
  \caption{Using CGP to play Atari. Red, green, blue pixel matrices are input to
    the evolved program, and evolved outputs determine the final controller
    action. Here, all legal controller actions are represented, but most games
    only use a subset of possible actions. Actions with a red mark indicate a
    button press.}
  \label{fig:scheme}
\end{figure*}

\subsection{Mixed Types}

In this work, the program inputs are pixel values of the Atari screen and
program outputs must be scalar values, representing the preference for a
specific action. Intermediate program nodes can therefore receive a mix of
matrix and scalar inputs. To handle this, each node's function was overloaded
with four possible input types: $(x, y), (x, \vec{y}), (\vec{x}, y), (\vec{x},
\vec{y})$. For some functions, broadcasting was used to apply the same function
to the scalar and matrix input types. In other functions, arity made it possible
to ignore the type of the $y$ argument. Some functions, however, such as
$std(\vec{x})$, require matrix input. In these cases, scalar $x$ input was
passed directly to output; in other words, these functions operated as a wire
when not receiving matrix input. In other functions, scalar input of either $x$
or $y$ is necessary. In these cases, the average value of matrix input is used.
Finally, some functions use inputs to index into matrices; when floating point
values are used to index into matrices, they are first multiplied by the number
of elements in the matrix and then rounded down.


To account for matrix inputs of different sizes, the minimum of each dimension
between the two matrices is taken. This inherently places more import on the
earlier values along each dimension than later ones, as the later ones will
often be discarded. However, between minimizing the sizes of the two matrices
and maximizing them, minimizing was found to be superior. Maximization requires
a padding value to fill in smaller dimensions, for which 0, 1, and $p_n$ were
used, but the resultant graphs were found to be highly dependent on this padding
value.

All functions in the chosen set are designed to operate over the domain $[-1.0,
  1.0]$. However, some functions, such as $std(\vec{x})$, return values outside
of this domain or are undefined for some values in this domain. Outputs are
therefore constrained to $[-1.0, 1.0]$ and \texttt{NaN} and $\inf$ values are
replaced with 0. This constraining operator is applied element-wise for matrix
output. While this appears to limit the utility of certain functions, there have
been instances of exaptation where evolution has used such functions with
out-of-domain values to achieve constant 0.0 or $p_n$ output.

The function set used in this work was designed to be as simple as possible
while still allowing for necessary pixel input processing. No image processing
library was used, but certain matrix functions allow for pixel input to inform
program output actions. The function set used in this work defined in tables
\autoref{tab:functions} and \autoref{tab:list_functions}. It is a large function
set and it is the intention of future work to find the minimal necessary
function set for Atari playing.

To determine the action taken, each node specified by an output gene is
examined. For nodes with output in matrix format, the average value is taken,
and for nodes with scalar output, the scalar value is taken. These output values
are then compared and the maximum value triggers the corresponding action.

\subsection{ALE}

In the ALE, there are 18 legal actions, corresponding to directional movements
of the controller (8), button pressing (1), no action (1), and controller
movement while button pressing (8). Not all games use every possible action;
some use as few as 4 actions. In this work, outputs of the evolved program
correspond only to the possible actions for each game. The output with the
highest value is chosen as the controller action.

An important parameter in Atari playing is frame skip \cite{braylan2000frame}.
In this work, the same frame skip parameter as in
\citet{hausknecht2014neuroevolution}, \citet{kelly2017emergent} and
\citet{mnih2015human} is used. Frames are randomly skipped with probability
$p_{fskip}=0.25$ and the previous controller action is replayed. This default
value was chosen as the highest value for which human play-testers were unable
to detect a delay or control lag \cite{machado17arcade}. This allows the results
from artificial controllers to be directly compared to human performance.

The screen representation used in this work is pixel values separated into red,
green, and blue layers. A representation of the full CGP and Atari scheme is
included in \autoref{fig:scheme}.

CGP parameter optimization was performed on a subset of the full game set
consisting of Boxing, Centipede, Demon Attack, Enduro, Freeway, Kung Fu Master,
Space Invader, Riverraid, and Pong. These games were chosen to represent a
variety of game strategies and input types. Games were played until completion
or until reaching 18000 frames, not including skipped frames.

\section{Results}
\label{sec:results}


\begin{figure}[h]
  \includegraphics[width=0.2\textwidth]{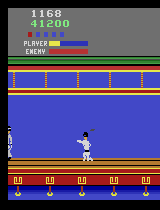}
  \includegraphics[width=0.2\textwidth]{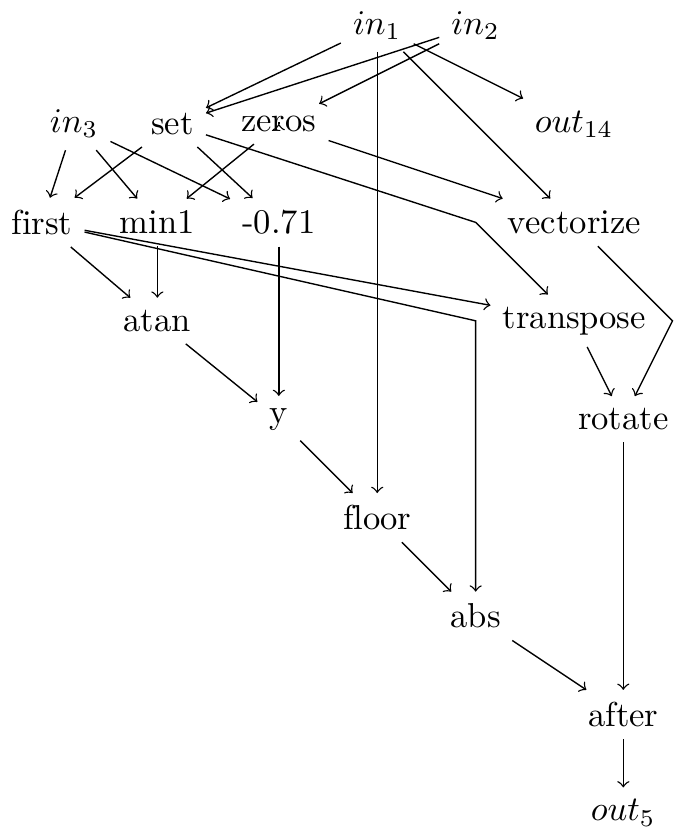}
  \caption{The Kung-Fu Master crouching approach and the functional graph of the
    player. Outputs which are never activated, and the computational graph
    leading to them, are omitted for clarity.}
  \label{fig:kungfu}
\end{figure}

By inspecting the resultant functional graphs of an evolved CGP player and
observing the node output values during its use, the strategy encoded by the
program can be understood. For some of the best performing games for CGP, these
strategies can remain incredibly simple. One example is Kung-Fu Master, shown in
\autoref{fig:kungfu}. The strategy, which can receive a score of 57800, is to
alternate between the crouching punch action (output 14), and a lateral movement
(output 5). The input conditions leading to these actions can be determined
through a study of the output program, but output 14 is selected in most cases
based simply on the average pixel value of input 1.

While this strategy is difficult to replicate by hand, due to the use of lateral
movement, interested readers are encouraged to try simply repeating the
crouching punch action on the Stella Atari emulator. The lateral movement allows
the Kung-Fu Master to sometimes dodge melee attacks, but the crouching punch is
sufficient to wipe out the enemies and dodge half of the bullets. In fact, in
comparison to the other attack options (low kick and high kick) it appears
optimal due to the reduced exposure from crouching. For the author, employing
this strategy by hand achieved a better score than playing the game normally,
and the author now uses crouching punches exclusively when attacking in this
game.

\begin{figure}[h]
  \includegraphics[width=0.2\textwidth]{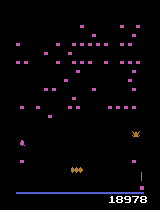}
  \includegraphics[width=0.2\textwidth]{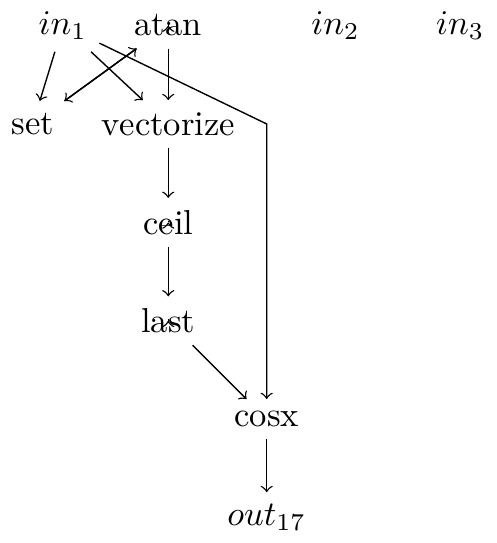}
  \caption{The Centipede player, which only activates output 17,
    down-left-and-fire. All other outputs are linked to null or constant zero
    inputs and are not shown.}
  \label{fig:centipede}
\end{figure}

Other games follow a similar theme. Just as crouching is the safest position in
Kung-Fu Master, the bottom left corner is safe from most enemies in Centipede.
The graph of an individual from early in evolution, shown in
\autoref{fig:centipede}, demonstrates this. While this strategy alone receives a
high score, it does not use any pixel input. Instead, output 17 is the only
active output, and is therefore repeated continuously. This action,
down-left-and-fire, navigates the player to the bottom left corner and
repeatedly fires on enemies. Further evolved individuals do use input to dodge
incoming enemies, but most revert to this basic strategy once the enemy is
avoided.

The common link between these simple strategies is that they are, on average,
effective. Evolution rewards agents by selecting them based on their overall
performance in the game, not based on any individual action. The policy which
the agent represents will therefore tend towards actions which, on average, give
very good rewards. As can be seen in the case of the Kung-Fu Master, which has
different attack types, the best of these is chosen. Crouching punch will
minimize damage to the player, maximizing the game's score and therefore the
evolutionary fitness. The policy encoded by the program doesn't incorporate
other actions because the average reward return for these actions is lower. The
safe locations found in these games can also be seen as an average maximum over
the entire game space; the players don't move into different positions because
those positions represent a higher average risk and therefore a worse
evolutionary fitness.

\begin{figure}[h]
  \includegraphics[width=0.2\textwidth]{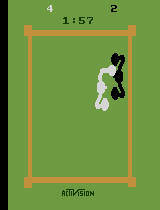}
  \caption{Boxing, a game that uses pixel input to continuously move and take
    different actions. Here, the CGP player has pinned the Atari player against
    the ropes by slowly advancing on it with a series of jabs.}
  \label{fig:boxing}
\end{figure}


Not all CGP agents follow this pattern, however. A counter example is boxing,
which pits the agent against an Atari AI in a boxing match. The CGP agent is
successful at trapping the Atari player against the ropes, leading to a quick
victory, as shown in \autoref{fig:boxing}. Doing this requires a responsive
program that reacts to the Atari AI sprite, moving and placing punches correctly
to back it into a corner. While the corresponding program can be read as a CGP
program, it is more complex and performs more input manipulation than the
previous examples. Videos of these strategies are included as supplementary
material.



Finally, in \autoref{tab:state}, CGP is compared to other state of the art
results. CGP performs better than all other compared artificial agents on 8
games, and is tied for best with HyperNEAT for one game. On a number of games
where CGP does not perform the best, it still achieves competitive scores to
other methods. However, there are certain games where CGP does not perform well.
There appears to be a degree of similarity between the evolved agents (TPG
\cite{kelly2017emergent}, HyperNEAT \cite{hausknecht2014neuroevolution}). There
is also a degree of similarity between the deep learning agents (Double
\cite{van2016deep}, Dueling \cite{wang2015dueling}, Prioritized
\cite{schaul2015prioritized}, and A3C \cite{mnih2016asynchronous}). The authors
attribute this similarity to the creation of a policy model for deep learning
agents, which is trained over a number of frames, as opposed to a player which
is evaluated over an entire episode, as is the case for the evolutionary
methods. This difference is discussed further in the next section.

\begin{table*}[t]
  {\small
  \begin{tabular}{l l l l l l l l l l l}
 & Human & Double &  Dueling &  Prioritized &  A3C FF &  A3C LSTM & TPG & HyperNEAT & CGP\\\hline
Alien & 6875 & 1033.4 & 1486.5 & 900.5 & 518.4 & 945.3 & \textbf{3382.7} & 1586 & 1978 {\tiny($\pm$ 268)}\\
Amidar & 1676 & 169.1 & 172.7 & 218.4 & 263.9 & 173 & \textbf{398.4} & 184.4 & 199 {\tiny($\pm$ 1)}\\
Assault & 1496 & 6060.8 & 3994.8 & 7748.5 & 5474.9 & \textbf{14497.9} & 2400 & 912.6 & 890.4 {\tiny($\pm$ 255)}\\
Asterix & 8503 & 16837 & 15840 & \textbf{31907.5} & 22140.5 & 17244.5 &  & 2340 & 1880 {\tiny($\pm$ 57)}\\
Asteroids & 13157 & 1193.2 & 2035.4 & 1654 & 4474.5 & 5093.1 & 3050.7 & 1694 & \textbf{9412} {\tiny($\pm$ 1818)}\\
Atlantis & 29028 & 319688 & 445360 & 593642 & \textbf{911091} & 875822 &  & 61260 & 99240 {\tiny($\pm$ 5864)}\\
Bank Heist & 734.4 & 886 & \textbf{1129.3} & 816.8 & 970.1 & 932.8 & 1051 & 214 & 148 {\tiny($\pm$ 18)}\\
Battle Zone & 3800 & 24740 & 31320 & 29100 & 12950 & 20760 & \textbf{47233.4} & 36200 & 34200 {\tiny($\pm$ 5848)}\\
Beam Rider & 5775 & 17417.2 & 14591.3 & \textbf{26172.7} & 22707.9 & 24622.2 &  & 1412.8 & 1341.6 {\tiny($\pm$ 21)}\\
Berzerk &  & 1011.1 & 910.6 & 1165.6 & 817.9 & 862.2 &  & \textbf{1394} & 1138 {\tiny($\pm$ 185)}\\
Bowling & 154.8 & 69.6 & 65.7 & 65.8 & 35.1 & 41.8 & \textbf{223.7} & 135.8 & 85.8 {\tiny($\pm$ 15)}\\
Boxing & 4.3 & 73.5 & \textbf{77.3} & 68.6 & 59.8 & 37.3 &  & 16.4 & 38.4 {\tiny($\pm$ 4)}\\
Breakout & 31.8 & 368.9 & 411.6 & 371.6 & 681.9 & \textbf{766.8} &  & 2.8 & 13.2 {\tiny($\pm$ 2)}\\
Centipede & 11963 & 3853.5 & 4881 & 3421.9 & 3755.8 & 1997 & \textbf{34731.7} & 25275.2 & 24708 {\tiny($\pm$ 2577)}\\
Chopper Comman & 9882 & 3495 & 3784 & 6604 & 7021 & \textbf{10150} & 7010 & 3960 & 3580 {\tiny($\pm$ 179)}\\
Crazy Climber & 35411 & 113782 & 124566 & 131086 & 112646 & \textbf{138518} &  & 0 & 12900 {\tiny($\pm$ 6620)}\\
Defender &  & 27510 & 33996 & 21093.5 & 56533 & 233021.5 &  & 14620 & \textbf{993010} {\tiny($\pm$ 2739)}\\
Demon Attack & 3401 & 69803.4 & 56322.8 & 73185.8 & 113308.4 & \textbf{115201.9} &  & 3590 & 2387 {\tiny($\pm$ 558)}\\
Double Dunk & -15.5 & -0.3 & -0.8 & \textbf{2.7} & -0.1 & 0.1 & 2 & 2 & 2 {\tiny($\pm$ 0)}\\
Enduro & 309.6 & 1216.6 & \textbf{2077.4} & 1884.4 & -82.5 & -82.5 &  & 93.6 & 56.8 {\tiny($\pm$ 7)}\\
Fishing Derby & 5.5 & 3.2 & -4.1 & 9.2 & 18.8 & \textbf{22.6} &  & -49.8 & -51 {\tiny($\pm$ 10)}\\
Freeway & 29.6 & 28.8 & 0.2 & 27.9 & 0.1 & 0.1 &  & \textbf{29} & 28.2 {\tiny($\pm$ 0)}\\
Frostbite & 4335 & 1448.1 & 2332.4 & 2930.2 & 190.5 & 197.6 & \textbf{8144.4} & 2260 & 782 {\tiny($\pm$ 795)}\\
Gopher & 2321 & 15253 & 20051.4 & \textbf{57783.8} & 10022.8 & 17106.8 &  & 364 & 1696 {\tiny($\pm$ 308)}\\
Gravitar & 2672 & 200.5 & 297 & 218 & 303.5 & 320 & 786.7 & 370 & \textbf{2350} {\tiny($\pm$ 50)}\\
H.E.R.O. & 25763 & 14892.5 & 15207.9 & 20506.4 & \textbf{32464.1} & 28889.5 &  & 5090 & 2974 {\tiny($\pm$ 9)}\\
Ice Hockey & 0.9 & -2.5 & -1.3 & -1 & -2.8 & -1.7 &  & \textbf{10.6} & 4 {\tiny($\pm$ 0)}\\
James Bond & 406.7 & 573 & 835.5 & 3511.5 & 541 & 613 &  & 5660 & \textbf{6130} {\tiny($\pm$ 3183)}\\
Kangaroo & 3035 & \textbf{11204} & 10334 & 10241 & 94 & 125 &  & 800 & 1400 {\tiny($\pm$ 0)}\\
Krull & 2395 & 6796.1 & 8051.6 & 7406.5 & 5560 & 5911.4 &  & \textbf{12601.4} & 9086.8 {\tiny($\pm$ 1328)}\\
Kung-Fu Master & 22736 & 30207 & 24288 & 31244 & 28819 & 40835 &  & 7720 & \textbf{57400} {\tiny($\pm$ 1364)}\\
Montezuma’s Revenge & 4367 & 42 & 22 & 13 & \textbf{67} & 41 & 0 & 0 & 0 {\tiny($\pm$ 0)}\\
Ms. Pacman & 15693 & 1241.3 & 2250.6 & 1824.6 & 653.7 & 850.7 & \textbf{5156} & 3408 & 2568 {\tiny($\pm$ 724)}\\
Name This Game & 4076 & 8960.3 & 11185.1 & 11836.1 & 10476.1 & \textbf{12093.7} &  & 6742 & 3696 {\tiny($\pm$ 445)}\\
Phoenix &  & 12366.5 & 20410.5 & 27430.1 & 52894.1 & \textbf{74786.7} &  & 1762 & 7520 {\tiny($\pm$ 1060)}\\
Pit Fall &  & -186.7 & -46.9 & -14.8 & -78.5 & -135.7 &  & \textbf{0} & \textbf{0} {\tiny($\pm$ 0)}\\
Pong & 9.3 & 19.1 & 18.8 & 18.9 & 5.6 & 10.7 &  & -17.4 & \textbf{20} {\tiny($\pm$ 0)}\\
Private Eye & 69571 & -575.5 & 292.6 & 179 & 206.9 & 421.1 & \textbf{15028.3} & 10747.4 & 12702.2 {\tiny($\pm$ 4337)}\\
Q*Bert & 13455 & 11020.8 & 14175.8 & 11277 & 15148.8 & \textbf{21307.5} &  & 695 & 770 {\tiny($\pm$ 94)}\\
River Raid & 13513 & 10838.4 & 16569.4 & \textbf{18184.4} & 12201.8 & 6591.9 & 3884.7 & 2616 & 2914 {\tiny($\pm$ 90)}\\
Road Runner & 7845 & 43156 & 58549 & 56990 & 34216 & \textbf{73949} &  & 3220 & 8960 {\tiny($\pm$ 2255)}\\
Robotank & 11.9 & 59.1 & \textbf{62} & 55.4 & 32.8 & 2.6 &  & 43.8 & 24.2 {\tiny($\pm$ 1)}\\
Seaquest & 20182 & 14498 & 37361.6 & \textbf{39096.7} & 2355.4 & 1326.1 & 1368 & 716 & 724 {\tiny($\pm$ 26)}\\
Skiing &  & -11490.4 & -11928 & -10852.8 & -10911.1 & -14863.8 &  & \textbf{-7983.6} & -9011 {\tiny($\pm$ 0)}\\
Solaris &  & 810 & 1768.4 & 2238.2 & 1956 & 1936.4 &  & 160 & \textbf{8324} {\tiny($\pm$ 2250)}\\
Space Invaders & 1652 & 2628.7 & 5993.1 & 9063 & 15730.5 & \textbf{23846} &  & 1251 & 1001 {\tiny($\pm$ 25)}\\
Star Gunner & 10250 & 58365 & 90804 & 51959 & 138218 & \textbf{164766} &  & 2720 & 2320 {\tiny($\pm$ 303)}\\
Tennis & -8.9 & -7.8 & \textbf{4.4} & -2 & -6.3 & -6.4 &  & 0 & 0 {\tiny($\pm$ 0)}\\
Time Pilot & 5925 & 6608 & 6601 & 7448 & 12679 & \textbf{27202} &  & 7340 & 12040 {\tiny($\pm$ 358)}\\
Tutankham & 167.6 & 92.2 & 48 & 33.6 & \textbf{156.3} & 144.2 &  & 23.6 & 0 {\tiny($\pm$ 0)}\\
Up n Down & 9082 & 19086.9 & 24759.2 & 29443.7 & 74705.7 & \textbf{105728.7} &  & 43734 & 14524 {\tiny($\pm$ 5198)}\\
Venture & 1188 & 21 & 200 & 244 & 23 & 25 & \textbf{576.7} & 0 & 0 {\tiny($\pm$ 0)}\\
Video Pinball & 17298 & 367823.7 & 110976.2 & 374886.9 & 331628.1 & \textbf{470310.5} &  & 0 & 33752.4 {\tiny($\pm$ 6909)}\\
Wizard of Wor & 4757 & 6201 & 7054 & 7451 & 17244 & \textbf{18082} & 5196.7 & 3360 & 3820 {\tiny($\pm$ 614)}\\
Yars Revenge &  & 6270.6 & 25976.5 & 5965.1 & 7157.5 & 5615.5 &  & 24096.4 & \textbf{28838.2} {\tiny($\pm$ 2903)}\\
Zaxxon & 9173 & 8593 & 10164 & 9501 & \textbf{24622} & 23519 & 6233.4 & 3000 & 2980 {\tiny($\pm$ 879)}\\
  \end{tabular}
  }
  \caption{Average CGP scores from five $1+\lambda$ evolutionary runs, compared
    to state of the art methods. Bold indicates the best score from an
    artificial player. Reported methods Double \cite{van2016deep}, Dueling
    \cite{wang2015dueling}, Prioritized \cite{schaul2015prioritized}, A3C
    \cite{mnih2016asynchronous}, TPG \cite{kelly2017emergent}, and HyperNEAT
    \cite{hausknecht2014neuroevolution} were chosen based on use of pixel input.
    Professional human game tester scores are from \citet{mnih2013playing}.}
  \label{tab:state}
\end{table*}

\section{Discussion}
\label{sec:discussion}

Taking all of the scores achieved by CGP into account, the capability of CGP to
evolve competitive Atari agents is clear. In this work, we have demonstrated how
pixel input can be processed by an evolved program to achieve, on certain games,
human level results. Using a function set based on list processing, mathematics,
and statistics, the pixel input can be properly processed to inform a policy
which makes intelligent game decisions.

The simplicity of some of the resultant programs, however, can be disconcerting,
even in the face of their impressive results. Agents like a Kung-Fu Master that
repeatedly crouches and punches, or a Centipede blaster that hides in the corner
and fires on every frame, do not seem as if they have learned about the game.
Even worse, some of these strategies do not use their pixel input to inform
their final strategies, a point that was also noted in
\citet{hausknecht2014neuroevolution}.

This is a clear demonstration of a key difficulty in evolutionary reinforcement
learning. By using the reward over the entire sequence as evolutionary fitness,
complex policies can be overtaken by simple polices that receive a higher
average reward in evolution. While CGP showed its capability to creating complex
policies, on certain games, there are more beneficial simple strategies which
dominate evolution. These simple strategies create local optima which can
deceive evolution.

In future work, the authors intend to use novelty metrics to encourage a variety
of policies. Novelty metrics have shown the ability to aid evolution in escaping
local optima. \cite{lehman2008exploiting} Furthermore, deep reinforcement
learning has shown that certain frames can be more important in forming the
policy than others \cite{schaul2015prioritized}. Similarly, evolutionary fitness
could be constrained to reward from certain frames or actions and not others.
Finally, reducing the frame count in evolution could also decrease the
computational load of evolving on the Atari set, as the same frame, action pairs
are often computed multiple times by similar individuals.

A more thorough comparison between methods on the Atari games is also necessary
as future work. Deep learning methods use frame counts, instead of episode
counts, to mark the training experience of a model. While the use of frame
skipping is consistent between all compared works, the random seeding of
environments and resulting statistical comparisons are difficult. The most
available comparison baseline is with published results, but these are often
averages or sometimes single episode scores. Finally, a thorough computational
performance comparison is necessary. The authors believe that CGP can achieve
the reported results much faster than other methods using comparable hardware,
as the main computational cost is performing the Atari games, but a more
thorough analysis is necessary.

In conclusion, this work represents a first use of CGP in the Atari domain, and
the first case of a GP method using pure pixel input. CGP was best among or
competitive with other artificial agents while offering agents that are far less
complex and can be read as a program. It was also competitive with human results
on a number of games and gives insight into better human playing strategies.
While there are many avenues for improvement, this work demonstrates that CGP is
competitive in the Atari domain.

\begin{acks}
This work is supported by ANR-11-LABX-0040-CIMI, within programme
ANR-11-IDEX-0002-02. This work was performed using HPC resources from CALMIP
(Grant P16043).
\end{acks}

\bibliographystyle{ACM-Reference-Format}
\bibliography{main} 

\end{document}